# SCU-CGAN: Enhancing Fire Detection through Synthetic Fire Image Generation and Dataset Augmentation[*]

Ju-Young Kim[1], Ji-Hong Park[1], Gun-Woo Kim[1†]
[1] School of Computer Science and Engineering,
Gyeongsang National University, Jinju, Republic of Korea
`wndudwkd003@gnu.ac.kr, hong_0002@gnu.ac.kr, gunwoo.kim@gnu.ac.kr`

**Abstract**

Fire has long been linked to human life, causing severe disasters and losses. Early detection is crucial, and with the rise of home IoT technologies, household fire detection systems have emerged. However, the lack of sufficient fire datasets limits the performance of detection models. We propose the SCU-CGAN model, which integrates U-Net, CBAM, and an additional discriminator to generate realistic fire images from non-fire images. We evaluate the image quality and confirm that SCU-CGAN outperforms existing models. Specifically, SCU-CGAN achieved a 41.5% improvement in KID score compared to CycleGAN, demonstrating the superior quality of the generated fire images. Furthermore, experiments demonstrate that the augmented dataset significantly improves the accuracy of fire detection models without altering their structure. For the YOLOv5 nano model, the most notable improvement was observed in the mAP@0.5:0.95 metric, which increased by 56.5%, highlighting the effectiveness of the proposed approach.

## 1 Introduction

Fire has been a crucial element for human survival, playing a significant role not only in enhancing survival capabilities but also in advancing societal culture and technology beyond its practical use. Unfortunately, uncontrolled fires have led to catastrophic disasters, resulting in severe damage and loss. Recently, models for detecting fires using CCTV footage have been actively researched, and the integration of home IoT technologies has led to an increase in the installation of small-scale CCTV systems in households (Vijayalakshmi, 2017), (Dilshad, 2023). However, as fire detection models become more complex, they require large-scale datasets to perform effectively. Especially, fire datasets are challenging to construct, and insufficient data relative to the model's complexity can lead to overfitting, which may hinder effective learning and reduce the model's generalization performance.

In the field of image generation research, dataset augmentation using GANs (Generative Adversarial Networks) has been actively explored (Lee, 2022). However, traditional GANs are prone to unstable training, leading to issues such as mode collapse and visual artifacts. To address these challenges, various modified models, such as DCGAN (Deep Convolutional GAN) and LSGAN (Least Squares GAN), have been proposed. Nevertheless, due to the ambiguous and indistinct visual characteristics of fire, generating fire regions that blend naturally with the background remains a difficult task. Additionally, image-to-image translation models like Pix2Pix and CycleGAN have been introduced,

---

with studies focusing on augmenting fire datasets using CycleGAN (Lee, 2022). However, Pix2Pix requires paired datasets, and CycleGAN's ambiguous mapping makes it challenging to preserve the fine details between two domains.

In this paper, we propose the SCU-CGAN (Six-CBAM with U-Net CycleGAN) model, which overcomes the limitations of the traditional CycleGAN by replacing the encoder-decoder structure with U-Net (Ronneberger, 2015) and incorporating CBAM (Woo, 2018) into the generator along with an additional discriminator for fire region generation. This architecture designates a specific region in a non-fire image as the fire generation area, focusing on that area to create natural fire images. Moreover, this structure significantly reduces the time and resources required for the labeling process, such as setting bounding boxes.

The main contributions of this paper are as follows:

1) **Proposal of the SCU-CGAN model**: We propose the SCU-CGAN model, which modifies the generator to be based on U-Net and CBAM, and adds an additional discriminator to generate fire images in specific target regions. The quality of the generated images was evaluated using FID, KID, Perceptual, and IoU metrics, with a separate analysis conducted for the generation quality of fire and background regions.
2) **Dataset augmentation and detection performance improvement**: The generated fire images were used to augment and strengthen the existing dataset for fire detection models, leading to improved accuracy as the models were retrained with the enhanced dataset. The improvement in accuracy was observed across all YOLOv5 model variants (nano, small, and large), demonstrating the effectiveness of the augmented dataset for different model sizes.

## 2  Related Works

### 2.1. CNN-based Lightweight Fire Detection Models

Fire detection is crucial for minimizing loss of life and property, and with advances in deep learning, research using convolutional neural networks (CNNs) has become active. Lightweight networks, such as YOLO (Redmon, 2016), have gained attention for enabling real-time detection in environments where large-scale servers are impractical. Additionally, models like CenterNet, which uses lightweight backbones like MobileNet (Howard, 2017), (Duan, 2019), predict object center points in an anchor-free approach. While these models are suitable for embedded systems, they may lack the expressiveness and accuracy of more complex models, requiring performance improvements through dataset augmentation and efficient training techniques.

### 2.2. GAN-based Image Generation Models

GAN (Generative Adversarial Network) is a model that generates realistic images through the competitive training of a generator and a discriminator, and it has garnered significant attention in the field of image generation. However, traditional GANs suffer from issues such as unstable training and mode collapse. To address these limitations, various modified models have been proposed. DCGAN (Radford, 2015), which replaces the MLP structure with CNN, and LSGAN (Mao, 2017), which uses the least squares error (MSE) as the loss function, are examples of such models. Additionally, models applying GAN to image translation tasks have emerged. Pix2Pix (Isola, 2017), for instance, uses a U-Net structure to preserve fine details, achieving good performance in translating images between domains. However, it requires paired datasets, which can be time-consuming and costly to construct. CycleGAN (Zhu, 2017) introduces cycle consistency loss, enabling unsupervised domain translation without paired datasets. While CycleGAN effectively captures the overall style of the translated images, the lack of explicit correspondences during training can lead to distortions in the structure of the original images. Therefore, additional structures are necessary to accurately reproduce the complex shapes and fine details of fire in fire image generation.

## 3 Proposed Method

The SCU-CGAN (Six-CBAM with U-Net CycleGAN) model proposed in this paper is based on the CycleGAN training method, as shown in Figure 1, and it learns both non-fire-to-fire and fire-to-non-fire transformations. The primary modification, as shown in Figure 2, is the incorporation of U-Net into the encoder-decoder structure of the CycleGAN generator, with CBAM applied to both the encoding-decoding process and skip connections. Additionally, the fire image generation process in this model utilizes a partially pre-generated Flame Patch, which is created by a separate LSGAN model. This design enables the SCU-CGAN generator to more easily and quickly generate fire regions. The LSGAN used to create the Flame Patch is not trained during this process; instead, noise is added to the fire generation region of the non-fire image, and the Flame Patch is applied at that location, resulting in a 6-channel image input for the generator.

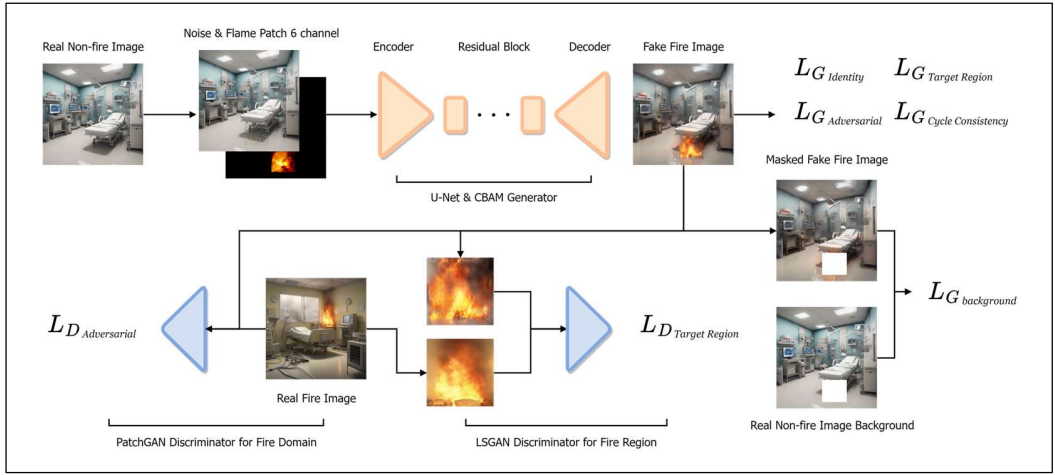

**Figure 1:** Flowchart for generating a fire image from a non-fire image in SCU-CGAN

### 3.1. SCU-CGAN Architecture

The model proposed in this paper employs two generators, $G_{F2NF}$ and $G_{NF2F}$, and four discriminators, $D_F, D_{NF}$ and $D_{FireRegion}, D_{NFireRegion}$ (hereafter referred to as $D_{FR}$ and $D_{NFR}$, respectively). $G_{F2NF}$ is responsible for transforming fire images into non-fire images, while $G_{NF2F}$ performs the reverse, converting non-fire images into fire images. Each generator is trained using the $L_{Adversarial}, L_{CycleConsistency}$ and $L_{Identitiy}$ loss functions to ensure the natural generation of fire-related features such as the fire atmosphere and burnt furniture through domain translation.

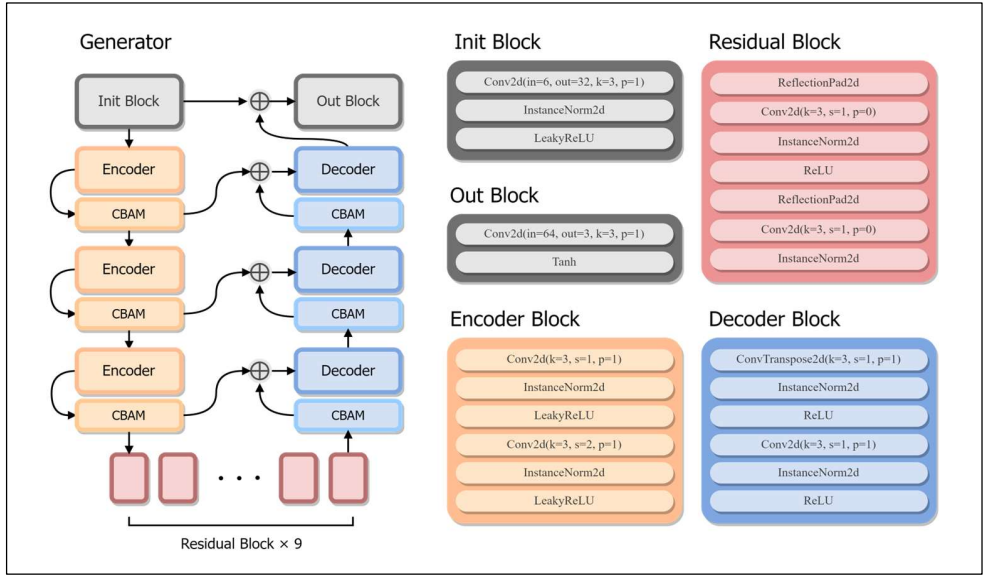

**Figure 2:** SCU-CGAN(Six-CBAM with U-Net CycleGAN) Generator architecture

The $L_{TargetRegion}$ loss is additionally applied to $G_{F2NF}, G_{NF2F}, D_{FR}$ and $D_{NFR}$. This loss function ensures that the generated fire images and the removed fire regions are appropriately generated for each domain. During fire domain transformation, $G_{NF2F}$ is trained to deceive $D_{FR}$ with the generated fire regions, while $D_{FR}$ learns to distinguish between real and generated fire images. Furthermore, $G_{F2NF}$ and $G_{NF2F}$ are also trained with the $L_{Background}$ loss, which helps restore the background area of the original image. $L_{Background}$ masks the target region of the original image, leaving only the background, compares it using $L_1$ loss. This process ensures that the background is restored similarly to the original, and both the fire and background regions are generated appropriately. In this study, $D_F$ and $D_{NF}$ use the PatchGAN discriminator employed in CycleGAN, while $D_{FR}$ and $D_{NFR}$ use the LSGAN discriminator. The $L_{TargetRegion}$ and $L_{Background}$ loss functions introduced in this model are as follows.

**Target Region Loss**: The formula 1 represents the adversarial loss function for the target region when transforming a non-fire image into a fire image. When performing the reverse transformation from a fire image to a non-fire image, the generator and discriminator are applied in the opposite manner.

$$\min_{G_{NF2F}} \max_{D_{FR}} L_{TargetRegion}(G_{NF2F}, D_{FR}) = \mathbb{E}_{x \sim P_{data}(x)}[\log D_{FR}(x)] + \mathbb{E}_{z \sim P_z(z)}\left[\log\left(1 - D_{FR}(G_{NF2F}(z))\right)\right] \quad (1)$$

**Background Loss**: The formula 2 represents the background loss, where M denotes the bounding box mask for the target region. Similarly, when transforming a fire image into a non-fire image, the process is applied in reverse.

$$L_{Background}(G_{NF2F}) = \mathbb{E}_{x \sim P_{data}(x)}[\|(1 - M) \odot G_{NF2F}(x) - (1 - M) \odot x\|_1] \quad (2)$$

## 3.2. CBAM and Skip Connection

CBAM (Convolutional Block Attention Module) is designed to enhance focus on important features

by combining channel and spatial attention mechanisms. In this model, CBAM is integrated between each encoder and decoder layer in the U-Net structure to emphasize critical features for fire region generation. The encoder utilizes CBAM to focus on significant regions within the background image and Flame Patch, extracting and compressing features. The features extracted by CBAM in the encoder are transmitted through skip connections, which are used to retain crucial details and highlight relevant spatial characteristics needed for realistic fire region generation.

In the decoder, CBAM emphasizes the regions where fire should be generated, creating a realistic fire image that is naturally synthesized with the background and Flame Patch. These skip connections allow the model to better preserve key features of the original image compared to CycleGAN. While CycleGAN's encoder-decoder structure struggles to retain detailed features, U-Net's skip connections help maintain features extracted from the encoder, allowing the model to restore the background to more closely resemble the original image. The formula used for the skip connections between the encoder and decoder is as follows.

**CBAM and Skip Connection**: The formula 3 represents the formula applied to the skip connections, where the CBAM-processed feature maps from the encoder and decoder are used. Here, 'd' refers to the decoder, 'e' to the encoder, 'n' indicates the corresponding layer, and 'F' denotes the feature map.

$$F_{d^n} = Decoder^n\left(concat(CBAM(F_{e^n}),\ CBAM(F_{d^{n-1}}))\right) \qquad (3)$$

## 4  Experimental Setup and Evaluation

### 4.1  Training Datasets

The images used for training and evaluation were generated using the Stable Diffusion XL model (Prodell, 2023), which is known for producing highly realistic images. Although relatively slow, the high-quality outputs were sufficient for training the fire-to-non-fire transformation models, effectively simulating both fire and non-fire scenarios in indoor environments. A total of 2,500 images, primarily of indoor hospital settings, were generated for training. To ensure accurate image generation, after filtering and labeling, two datasets of 1,000 images each were created, along with approximately 5,000 non-fire images for dataset augmentation testing.

To evaluate the performance of the proposed model, we used FID (Fréchet Inception Distance), KID (Kernel Inception Distance), Perceptual Loss, and IoU (Intersection over Union) as evaluation metrics.

| Model Name | FID | KID | Perceptual Loss | IoU |
| --- | --- | --- | --- | --- |
| LSGAN | 333.609 | 0.489 | 1.785 | - |
| CycleGAN | 101.413 | 0.136 | 1.774 | 0.372 |
| CycleGAN + U-Net | 75.37 | 0.087 | 1.754 | 0.174 |
| CycleGAN + U-Net + CBAM | 74.214 | 0.088 | 1.742 | 0.284 |
| CycleGAN + U-Net + $L_{BG}, L_{TR}$ | 79.973 | 0.077 | 1.727 | 0.726 |
| SCU-CGAN(ours) | **64.282** | **0.068** | **1.722** | **0.765** |

**Table 1:** Quantitative assessment of image quality

(Heusel, 2017), (Bińkowski, 2018), (Johnson, 2016), (Zhu, 2017) Lower FID and KID scores indicate a closer distribution between the real and generated images. Perceptual Loss compares high-level features of the images rather than simple pixel differences, with lower values indicating higher structural similarity. Finally, IoU measures the similarity between the predicted fire bounding boxes and the actual bounding boxes, with values closer to 1 indicating more accurate generation of the fire regions. All evaluation metrics used in this study were tested with pre-trained models.

## 4.2 Qualitative and Quantitative Experiments

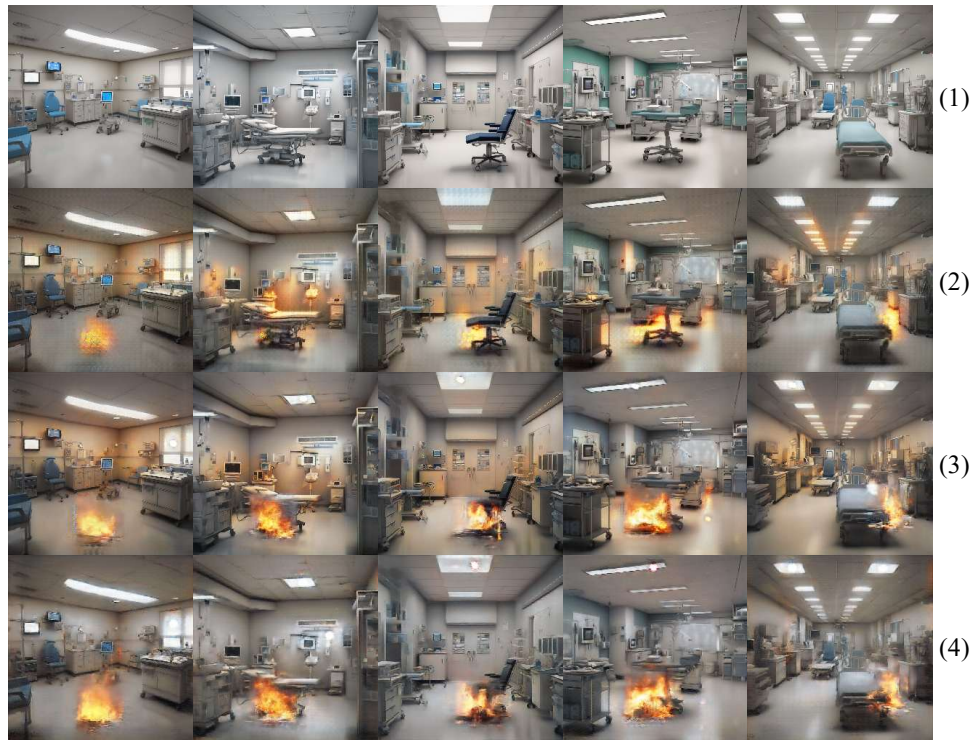

**Figure 3:** (1) Non-fire image, (2) CycleGAN(6ch), (3) U-Net + BG + TR (4) SCU-CGAN(Proposed)

In this study, both qualitative and quantitative evaluations were conducted to assess the image quality of the generative models. Table 1 compares the models, including LSGAN, CycleGAN, CycleGAN with U-Net and CBAM, and the proposed model, across various evaluation metrics. Figure 3 (1) shows a non-fire image used to generate a fire image, while (2) and (3) depict fire images generated by CycleGAN and CycleGAN with U-Net, background loss, and target region loss, respectively. In (2), the background contains noticeable noise, and the fire appears more like light beams. However, when U-Net was added to CycleGAN, although the overall image quality improved, the fire did not appear in the target region. This highlights why previous CycleGAN research did not adopt U-Net.

When CBAM was added to U-Net, a faint yet more fire-like image was generated. As shown in Table 1, the IoU increased when CBAM was applied, indicating that the model could focus on the target region and generate fire-like images. Furthermore, when target region and background loss functions were added to the U-Net structure, while the overall image quality decreased, the fire image became clearer. In Figure 3 (3), the fire generated is more distinct compared to (2), though the image quality is slightly reduced due to unnatural blending with the background.

In contrast, Figure 3 (4) shows that the proposed model generates more realistic fire and blends it seamlessly into the background, offering a superior visual result. The results show that the proposed model outperforms the others in terms of image generation quality across all metrics. This experiment confirmed that the additional discriminator and background loss function made the fire generation area appear more realistic, and that the skip connections of the CBAM-applied U-Net effectively captured finer details, generating fire images more realistically. This structure also showed that it could naturally generate the fire in relation to the background furniture and objects.

## 4.3 Evaluating Datasets for Augmented Training

In this study, we evaluated the effectiveness of augmenting 1,000 original fire images with 5,000 additional fire images generated by the SCU-CGAN model to enhance the performance of the YOLOv5 model. Table 2 shows the performance of the YOLOv5 model when trained on the original fire images compared to the augmented dataset. 'n', 's', and 'l' represent nano, small, and large models, respectively, with the large model having the most parameters among the three. The nano and large models showed the greatest improvement in performance, confirming the significance of dataset augmentation.

In terms of evaluation metrics, precision refers to the proportion of correctly identified fire regions out of all predicted fire regions, while recall measures the proportion of actual fire regions that were correctly identified by the model. mAP@0.5 represents the mean average precision at an Intersection over Union (IoU) threshold of 0.5, which indicates how well the predicted bounding boxes overlap with the ground truth boxes. The more stringent mAP@0.5:0.95 measures the average precision over multiple IoU thresholds, from 0.5 to 0.95, providing a more comprehensive evaluation of the model's performance.

**Precision**: The formula 4 represents precision, which refers to the proportion of correctly identified fire regions out of all predicted fire regions. In other words, it measures the accuracy of the model's positive predictions. Mathematically, precision is calculated as.

$$Precision = \frac{True\ Positive\ (TP)}{True\ Positives\ (TP) + False\ Positives\ (FP)} \quad (4)$$

**Recall**: The formula 5 represents recall, which measures the proportion of actual fire regions that were correctly identified by the model. It indicates the model's ability to detect fires when they occur. Mathematically, recall is calculated as.

$$Recall = \frac{True\ Positive\ (TP)}{True\ Positives\ (TP) + False\ Negatives\ (FN)} \quad (5)$$

| Model Name | Precision | Recall | mAP@0.5 | mAP@0.5:0.95 |
|---|---|---|---|---|
| YOLOv5n | 0.72 | 0.804 | 0.808 | 0.354 |
| YOLOv5n + Proposed | **0.923** | **0.911** | **0.951** | **0.554** |
| YOLOv5s | 0.854 | 0.889 | 0.914 | 0.529 |
| YOLOv5s + Proposed | **0.887** | **0.929** | **0.945** | **0.539** |
| YOLOv5l | 0.778 | 0.831 | 0.832 | 0.365 |
| YOLOv5l + Proposed | **0.904** | **0.902** | **0.924** | **0.559** |

**Table 2:** Compare training results before and after dataset augmentation

As shown in Table 2, dataset augmentation with SCU-CGAN-generated images resulted in considerable improvements across all models. The nano ('n') and large ('l') models, in particular, showed the most significant gains in precision, recall, and mAP metrics. For example, the nano model's precision increased from 0.72 to 0.923 and recall improved from 0.804 to 0.911, indicating that the model became more effective at accurately identifying fire regions. Higher precision means that, among the instances where the model predicted a fire, a greater proportion were actual fire cases, reflecting a reduction in false positives. Similarly, higher recall indicates that, among the actual fire cases, a greater proportion were correctly identified by the model, reducing the likelihood of missed detections (false negatives). This improvement in recall is particularly important in fire detection systems, as failing to detect an actual fire can lead to serious safety risks.

Furthermore, the nano model's mAP@0.5 increased from 0.808 to 0.951, and the large model improved from 0.832 to 0.924. Similarly, the mAP@0.5:0.95 values also saw substantial increases, further demonstrating the effectiveness of the augmented dataset in improving the model's ability to accurately detect and localize fire regions. These improvements in mAP scores reflect the enhanced overall accuracy of the models in correctly localizing fire regions within the images. Collectively, these results underscore the importance of dataset augmentation in enhancing model performance, especially for models with fewer or more extensive parameters.

# 5  Conclusion

In this study, we proposed a dataset augmentation model using SCU-CGAN to improve the performance of fire detection models. To overcome the limitations of the existing CycleGAN, we employed a structure combining U-Net and CBAM, enabling the natural generation of fire images focused on specific regions. Additionally, we introduced an auxiliary discriminator and loss functions to generate more realistic target regions, successfully producing more refined fire images. The model proposed in this paper demonstrated superior performance compared to LSGAN and CycleGAN, as measured by FID, KID, Perceptual Loss, and IoU metrics. Specifically, the proposed SCU-CGAN achieved a 41.5% improvement in KID over CycleGAN, highlighting the enhanced quality of generated images. Furthermore, dataset augmentation using SCU-CGAN resulted in a 56.5% increase in the mAP@0.5:0.95 metric for the YOLOv5 nano model, demonstrating the practical benefits of the augmented dataset in fire detection performance.

The results confirm that training with augmented datasets significantly improves the performance of fire detection models. As future work, we aim to construct and augment real-world fire datasets for both indoor and outdoor environments to further enhance the performance of fire detection models. We plan to compare various dataset augmentation techniques and experimentally evaluate the effectiveness of the proposed method in this study. Through this study, we hope to address the challenges of building fire datasets and contribute to improving the performance of lightweight object detection models.

# 6  Acknowledgements

This research was supported by Basic Science Research Program through the National Research Foundation of Korea (NRF), funded by the Ministry of Education, Science and Technology (NRF-2021R1G1A1006381), and "Leaders in Industry-university Cooperation 3.0" Project, supported by the Ministry of Education and National Research Foundation of Korea.

# References

Vijayalakshmi, S. R., & Muruganand, S. (2017, February). A survey of Internet of Things in fire detection and fire industries. In *2017 International Conference on I-SMAC (IoT in Social, Mobile, Analytics and Cloud)(I-SMAC)* (pp. 703-707). IEEE.

Dilshad, N., Khan, S. U., Alghamdi, N. S., Taleb, T., & Song, J. (2023). Towards efficient fire detection in IoT environment: a modified attention network and large-scale dataset. *IEEE Internet of Things Journal*.

Lee, H., Kang, S., & Chung, K. (2022). Robust data augmentation generative adversarial network for object detection. *Sensors*, *23*(1), 157.

Lee, H., Kang, S., & Chung, K. (2022, October). Object detection with dataset augmentation for fire images based on gan. In *2022 13th International Conference on Information and Communication Technology Convergence (ICTC)* (pp. 2118-2123). IEEE.

Ronneberger, O., Fischer, P., & Brox, T. (2015). U-net: Convolutional networks for biomedical image segmentation. In *Medical image computing and computer-assisted intervention–MICCAI 2015: 18th international conference, Munich, Germany, October 5-9, 2015, proceedings, part III 18* (pp. 234-241). Springer International Publishing.

Woo, S., Park, J., Lee, J. Y., & Kweon, I. S. (2018). Cbam: Convolutional block attention module. In *Proceedings of the European conference on computer vision (ECCV)* (pp. 3-19).

Redmon, J. (2016). You only look once: Unified, real-time object detection. In *Proceedings of the IEEE conference on computer vision and pattern recognition*.

Howard, A. G. (2017). Mobilenets: Efficient convolutional neural networks for mobile vision applications. *arXiv preprint arXiv:1704.04861*.

Duan, K., Bai, S., Xie, L., Qi, H., Huang, Q., & Tian, Q. (2019). Centernet: Keypoint triplets for object detection. In *Proceedings of the IEEE/CVF international conference on computer vision* (pp. 6569-6578).

Radford, A. (2015). Unsupervised representation learning with deep convolutional generative adversarial networks. *arXiv preprint arXiv:1511.06434*.

Mao, X., Li, Q., Xie, H., Lau, R. Y., Wang, Z., & Paul Smolley, S. (2017). Least squares generative adversarial networks. In *Proceedings of the IEEE international conference on computer vision* (pp. 2794-2802).

Isola, P., Zhu, J. Y., Zhou, T., & Efros, A. A. (2017). Image-to-image translation with conditional adversarial networks. In *Proceedings of the IEEE conference on computer vision and pattern recognition* (pp. 1125-1134).

Zhu, J. Y., Park, T., Isola, P., & Efros, A. A. (2017). Unpaired image-to-image translation using cycle-consistent adversarial networks. In *Proceedings of the IEEE international conference on computer vision* (pp. 2223-2232).

Podell, D., English, Z., Lacey, K., Blattmann, A., Dockhorn, T., Müller, J., ... & Rombach, R. (2023). Sdxl: Improving latent diffusion models for high-resolution image synthesis. *arXiv preprint arXiv:2307.01952*.

Heusel, M., Ramsauer, H., Unterthiner, T., Nessler, B., & Hochreiter, S. (2017). Gans trained by a two time-scale update rule converge to a local nash equilibrium. *Advances in neural information processing systems*, *30*.

Bińkowski, M., Sutherland, D. J., Arbel, M., & Gretton, A. (2018). Demystifying mmd gans. *arXiv preprint arXiv:1801.01401*.

Johnson, J., Alahi, A., & Fei-Fei, L. (2016). Perceptual losses for real-time style transfer and super-resolution. In *Computer Vision–ECCV 2016: 14th European Conference, Amsterdam, The Netherlands, October 11-14, 2016, Proceedings, Part II 14* (pp. 694-711). Springer International Publishing.